\documentclass[10pt]{article} % For LaTeX2e

\usepackage[preprint]{rlj}           % Should be uncommented for submission
\usepackage{amssymb}            % Defines common symbols like \mathbb R
\usepackage{mathtools}          % Extends amsmath, providing common math tools
\usepackage{mathrsfs}           % Enables \mathscr, which can work in cases that \mathcal does not
\usepackage{graphicx}           % For including images
\usepackage{subcaption}         % Allows for the use of subfigures and subcaptions
\usepackage[space]{grffile}     % For spaces in image names
\usepackage{url}                % For displaying URLs
\usepackage{lipsum}             % For placeholder text
\usepackage{times}
\usepackage{soul}
\usepackage[utf8]{inputenc}
\usepackage{caption}
\usepackage{amsthm}
\usepackage{amsfonts}       % blackboard math symbols, bold math
\usepackage{booktabs}
\usepackage{algorithm}
\usepackage{algorithmic}
\usepackage{tikz}
\usepackage{svg}
\usepackage{lipsum}  
\usepackage{multirow}

\newtheorem{definition}{Definition}

\renewcommand{\sectionautorefname}{Section}

\title{Zero-Incentive Dynamics: a look at reward sparsity through the lens of unrewarded subgoals}

\setrunningtitle{Zero-Incentive Dynamics}

\author{Yannick Molinghen\textsuperscript{1}, Tom Lenaerts\textsuperscript{1,2,3}}

\emails{yannick.molinghen@ulb.be, tom.lenaerts@ulb.be}

\affiliations{
$^{1}$\textbf{Machine Learning Group, Université Libre de Bruxelles, Belgium}\\
$^{2}$\textbf{AI Lab, Vrije Universiteit Brussel, Belgium}\\
$^{3}$\textbf{Center for Human-compatible AI, UC Berkley, USA}
}

\contribution{
    We show that reward sparsity is not a reliable metric for the difficulty of a task.
}
{
    Reward sparsity is often presented as the source of the difficulty of an RL task.
}

\contribution{
    We formally define Zero-Incentive Dynamics, the fact that the accomplishment of subtasks is not rewarded.
}{
    Zero-Incentive Dynamics (also referred to as Partially Ordered Subtasks) has already been identified in previous works but has only been defined intuitively.
}

\contribution{
    We show that state-of-the-art subgoal-oriented methods are unable to outperform more \textit{general-purpose} methods under Zero-Incentive Dynamics.
}{
    The presented methods have shown to outperform more \textit{general-purpose} algorithms in other tasks that do not have Zero-Incentive Dynamics.
}

\contribution{
    We show that the quality of the deep RL policy correlates with the distance (in time steps) between the achievement of a subtask and the collection of the corresponding reward.
}{
    None
}

 % Your keywords

\summary{    
    This work re-examines the commonly held assumption that the frequency of rewards is a reliable measure of task difficulty in reinforcement learning. We identify and formalize a structural challenge that undermines the effectiveness of current policy learning methods: when essential subgoals do not directly yield rewards. We characterize such settings as exhibiting zero-incentive dynamics, where transitions critical to success remain unrewarded. We show that state-of-the-art deep subgoal-based algorithms fail to leverage these dynamics and that learning performance is highly sensitive to the temporal proximity between subgoal completion and eventual reward. These findings reveal a fundamental limitation in current approaches and point to the need for mechanisms that can infer latent task structure without relying on immediate incentives.
}

\begin{document}

%\makeCover  % Create the cover page
\maketitle  % Make the title section

\begin{abstract}
    This work re-examines the commonly held assumption that the frequency of rewards is a reliable measure of task difficulty in reinforcement learning. We identify and formalize a structural challenge that undermines the effectiveness of current policy learning methods: when essential subgoals do not directly yield rewards. We characterize such settings as exhibiting zero-incentive dynamics, where transitions critical to success remain unrewarded. We show that state-of-the-art deep subgoal-based algorithms fail to leverage these dynamics and that learning performance is highly sensitive to the temporal proximity between subgoal completion and eventual reward. These findings reveal a fundamental limitation in current approaches and point to the need for mechanisms that can infer latent task structure without relying on immediate incentives.
\end{abstract}

\section{Introduction}
%In real-life, the outcome of many tasks can simply be expressed as a success or a failure, and measuring the progress of the task until completion can sometimes be difficult, even impossible. Playing chess or designing a new drug are examples of such tasks. 

In reinforcement learning \citep[RL]{sutton_reinforcement_2018}, agents learn an optimal policy by maximizing the expected cumulative reward through interactions with the environment. In cases where there is little-to-no feedback until the completion of the task, reward is said to be sparse. In such cases, learning becomes significantly more difficult due to the sparsity (even absence) of informative gradients for policy improvement \citep{ocana_overview_2023}. This challenge has motivated substantial research into mitigation techniques such as  reward shaping \citep{y_ng_policy_1999}, intrinsic motivation \citep{pathak_curiosity-driven_2017, burda_exploration_2018}, and hierarchical RL \citep{sutton_between_1999}.

While these approaches aim to compensate for sparse signals by injecting auxiliary rewards or by introducing structure, they often treat sparsity as a scalar property of the reward function, namely the frequency or density of non-zero rewards. However, this view neglects the structure of the problem and the location of the reward with regard to subgoals in the tasks.

In this work, we aim to shed light on why reward sparsity should not be considered to be the sole metric of a problem difficulty and look at reward sparsity through the lens of subtasks whose accomplishment is unrewarded. Two recent works \citep{ocana_overview_2023,molinghen_laser_2024} respectively introduced Partially Ordered Subtasks and Zero-Incentive Dynamics (ZID) intuitively to refer to essential transitions that are not directly reinforced despite being necessary for the task success.

We begin by empirically showing in \autoref{sec:sparsity-2} that reward sparsity alone is an insufficient indicator of task difficulty. Specifically, we construct a series of environments with increasing reward density where exploration actually becomes harder, demonstrating that the distribution and alignment of rewards with subtask transitions matters more than their global frequency in that example. Then, we formalize in \autoref{sec:zid-definitions} the notion of ZID as a structural property of the underlying Markov Decision Process \citep[MDP]{bellman_dynamic_1957} characterized by unrewarded but mandatory transitions that shape the state space in bottlenecks. Using a graph-theoretic perspective, we define ZID in terms of cut-sets over the transition graph of the MDP and distinguish it from general reward sparsity by focusing on the causal and temporal relationship between subgoal completion and eventual reward. Next, in \autoref{sec:subgoal-oriented} we provide evidence that current state-of-the-art subgoal-oriented methods \citep{sutton_between_1999,jeon_maser_2022,xu_haven_2023} fail to exploit the presence of subgoals when they have ZID. Despite their strong performance in other benchmarks, these methods perform no better than other RL algorithms under ZID. Finally, we investigate in \autoref{sec:experimental-investigation} the impact of reward delay on deep RL by introducing reward shaping with controlled delays between subgoal completion and reward delivery. We show that even when the reward density remains constant, decreasing this delay substantially improves the quality of the policy. This suggests that the proximity of rewards to subtask completions, rather than their mere presence, is critical for effective learning.

Our findings highlight a gap in the current RL paradigm: the lack of mechanisms to identify unrewarded subtasks. Addressing this gap may require new architectures or representations that can detect structural dependencies in the environment, even in the absence of immediate reward signals.

\section{Related work}

\subsection{Markov Decision Processes}
Markov Decision Process \cite[MDP]{bellman_dynamic_1957} provide a model of sequential decision-making under uncertainty and are the main formalism used in Reinforcement Learning \citep[RL]{sutton_reinforcement_2018}. An MDP $M$ are defined by the tuple $M=\left<S, A, R, T\right>$ where $S$ is the set of state (state space), $A$ is the set of actions (action space), $R: S \times A \times S \rightarrow \mathbb{R}$ is a reward function, and $T: S \times A \times S \rightarrow \left[0, 1\right]$ is the transition function that determines the probability of that in state $s$, taking action $a$ results in state $s'$. In particular, we note $S_0 \subset S$ the set of initial states and $S_G \subset S$ the set of goal states.\\
Given an MDP $M=\left<S, A, R, T\right>$, a directed weighted graph \citep{wilson_introduction_2009} $G=(V, E, W)$ can be constructed to represent $M$, where $V = S$ is the set of vertices, $E=\left\{ (s, a, s')~|~s, s' \in S,~a \in A\right\}$ is the set of edges, and $W = R$ is the weight function that associates to each edge the reward for taking the corresponding transition.

\subsection{Reward sparsity}
\label{sec:sparsity}
In recent years, the concept of reward sparsity has regularly been discussed in the field of deep RL and has generally been presented as the source of the difficulty of the problem under study~\citep{andrychowicz_hindsight_2017, burda_exploration_2018, ladosz_exploration_2022, trott_keeping_2019}. However, only a handful of these studies define reward sparsity, let alone a justify why reward sparsity is problematic. Regardless, a wide variety of methods have been introduced to cope with reward sparsity. \citet{meng_research_2024} splits them into three categories: reward shaping (or manual labelling), intrinsic motivation, and hierarchical RL.

\label{sec:reward-shaping}
\textbf{Reward shaping} is the most straightforward way to tackle reward sparsity and relies on the introduction of additional rewards based on domain knowledge. \citet{randlov_learning_1998} and \citet{amodei_concrete_2016}, among others, have illustrated how naive reward shaping can alter the definition of the task, resulting in a change to the optimal policy $\pi^*$ and causing the agent's policy to no longer fulfil the task designer's intended objective. This is why, as early as 1999, \citet{y_ng_policy_1999} introduce Potential-Based Reward Shaping (PBRS) that modifies the reward function $R$ as shown in \autoref{eq:pbrs}. PBRS relies on the definition of a potential function $\phi$ and ensures that the optimal policy $\pi^*$ remains unchanged. However, even though PBRS guarantees the invariance of $\pi^*$, this technique relies on domain knowledge, is problem-specific and can therefore not readily be generalised to any task. Furthermore, PBRS can be extremely challenging to implement in practice, if not impossible due to a lack of domain-knowledge.

\begin{equation}
    \label{eq:pbrs}
    R'(s, a, s') = R(s, a, s') + \gamma \phi(s) - \phi(s')
\end{equation}

\textbf{Exploration bonuses} aim at providing an extra reward to the agent according to the novelty of the state visited and can take various forms. In its most simplistic form, intrinsic curiosity is a statewise counter and the agent receives an extra reward that is annealed over the number of times this state has been visited \citep{tang_exploration_2017}. More complex approaches rely on the training of neural networks to predict an embedding of a state and use the prediction error as measure of novelty \citep{pathak_curiosity-driven_2017,burda_exploration_2018}. Although these methods have shown very good results in environments with sparse rewards, it it important to realise that they do not guarantee policy invariance and can therefore suffer the same problems as the ones discussed in \autoref{sec:reward-shaping}.

\label{sec:hrl}
\textbf{Hierarchical RL} is an approach whose objective is to identify intermediate states of interest that are referred to as subgoals \citep{sutton_between_1999} in order to decompose a large task in smaller and more manageable ones. The literature distinguishes two categories of subgoal-oriented problems, a former where subgoals are explicitly disclosed by the environment \citep{ahilan_feudal_2019,geng_hisoma_2024} and a latter where they are not \citep{xu_haven_2023,jeon_maser_2022}. Leveraging subgoals often involves hierarchical structures where a high-level agent, or meta-agent, trains a high-level policy $\pi^h$ to select subgoals for a low-level policy $\pi^l$ to achieve. This decomposition reduces the exploration space and enables the agent to focus on intermediate objectives, making it easier to solve long-term tasks, but also relies on the ability of the meta-agent to identify these subgoals.

\subsection{Partially Ordered Subtasks}
\label{sec:deep-rl-under-sparsity}
\citet{ocana_overview_2023} identify nine features of the underlying MDP that negatively or positively impact deep RL under reward sparsity. Amongst others, the presence of \textit{Partially Ordered Subtasks} (POS) is one of them. From a high-level point of view, the authors discuss that the presence of subtasks whose completion is not rewarded nor indicated by the environment constitutes an obstacle to deep RL. These subtasks are partially ordered in the sense that their completion can be performed in different orders but the agent has to complete them all to fulfil the overall task, which yields a positive feedback.

\citet{ocana_overview_2023} discuss POS with the example of a problem where the agent has two subtasks to complete: learn to swim and learn to climb. The authors explain that even if exploration bonuses (discussed in \autoref{sec:reward-shaping}) can have the agent learn how to swim initially, there is no guarantee that when the agent starts to learn to climb, it does not forget how to swim, which \citet{french_catastrophic_1999} refer to as catastrophic forgetting. Although the authors indeed discuss the theoretical implications of POS, they do not provide evidence of why their presence can negatively impact the learning process.

\subsection{Laser Learning Environment}
\label{sec:lle}

\begin{figure}
    \centering
    \includegraphics[width=0.3\linewidth]{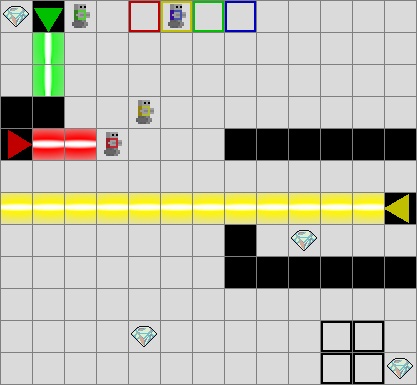}
    \caption{Laser Learning Environment. This map has four agents and three lasers that can be blocked. The red agent is currently blocking the red laser.}
    \label{fig:lle}
\end{figure}

The Laser Learning Environment \citep[LLE]{molinghen_laser_2024} illustrated in \autoref{fig:lle} has recently been introduced as a challenging MARL environment with sparse rewards. LLE is a deterministic fully observable cooperative grid-world in which agents can collect gems and must reach the exit represented by the black squares. The cooperative dynamics of LLE revolve around laser-blocking: agents can block lasers of the same colour as theirs, allowing other agents to pass, but die if they walk into a beam of a different colour. As such, collecting a gem or exiting the game is rewarded by +1, while dying in a laser is punished by -1 and terminates the episode.

\citet{molinghen_laser_2024} showed that state-of-the-art methods such as Value Decomposition Networks \citep[VDN]{sunehag_value-decomposition_2018} or QMIX \citep{rashid_qmix_2018} were unable to complete LLE and hypothesize that the fact that crossing lasers is not rewarded -- which they refer to as Zero-Incentive Dynamics (ZID) -- is the main cause of this failure but provided no evidence of their claim.

\section{Reward sparsity}
\label{sec:sparsity-2}
Although the methods mentioned in \autoref{sec:sparsity} have shown some success to mitigate reward sparsity in some tasks, we argue that there remain misconceptions about reward sparsity, notable as to whether or why it can be a challenge. In this section, we first provide a formal definition of reward sparsity and then show why reward sparsity alone is not a reliable indicator of the difficulty of a task.

\subsection{Definitions}
Intuitively, reward sparsity refers to the property of an MDP whose reward signal where most transitions yield the same base reward $r_b$, typically zero, and only a handful of transitions provide a reward superior to $r_b$, typically upon completing a goal or a task. 

\begin{definition}[Reward density]
    \label{def:density}
    Let $M = \left<S, A, R, T\right>$ be an MDP and $G_M = (V, E, W)$ be the directed weighted graph induced by $M$. Let $E^+$ be the set of edges with a meaningful reward, i.e. $E^+ = \left\{e \in E~|~W(e) > r_b\right\}$. The reward density of $M$ is $\mathcal{D}_M=\frac{|E^+|}{|E|}$.
\end{definition}

\begin{definition}[Reward sparsity]
    \label{def:sparsity}
    Let $M$ be an MDP. $M$ has sparse rewards if $0 < \mathcal{D}_M \ll 1$.
\end{definition}

As a result of definitions \ref{def:density} and \ref{def:sparsity}, a sparse MDP has a low reward density, and a dense MDP does not have sparse rewards. If these definitions do not provide a threshold from which a reward is sparse, they enable the comparison and ordering of MDPs with regard to reward sparsity.

\subsection{Reward sparsity, an incomplete metric}
\label{sec:denser}
\begin{figure}
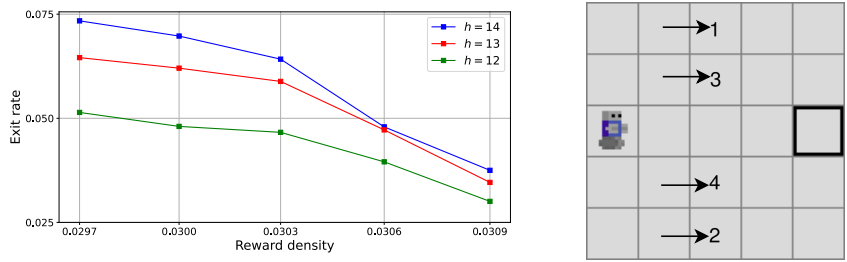

    \centering
    \begin{subfigure}{0.55\linewidth}
        \includesvg[height=0.15\textheight]{plots/exit_rate_per_density.svg}
    \end{subfigure}
    \begin{subfigure}{0.25\linewidth}
        \includesvg[height=0.15\textheight]{pictures/exploration.drawio.svg}
        \vfill
    \end{subfigure}
    \caption{(Left) Average exit rate over with random exploration for 200k steps for different time horizons $h$. Counter-intuitively, the exit rate decreases when the reward density increases. (Right) The map on which the left plot is based on. The arrows labelled with $1, \dots, 4$ indicate which state-actions are disabled for $M_1, \dots, M_4$.}
    \label{fig:density}
\end{figure}

We illustrate that reward sparsity is not a reliable indicator of the difficulty of a problem with the environment $M$ shown in \autoref{fig:density}. The underlying graph has 101 edges, 3 of which lead to the goal state in black and therefore yield a positive reward (details can be found in the Supplementary Materials~\ref{apx:density-computation}). As a result, $\mathcal{D}_M = \frac{3}{101} \approx 0.0297$.

Consider $M_1, \dots, M_4$, four variations of $M$ where $M_n$ is the same MDP as $M$ except that the transitions indicated by the arrows in \autoref{fig:density} whose label is $\leq n$ are disabled. For instance, $M_2$ corresponds to $M$ with the arrows labelled with $1$ and $2$ disabled. Hence, $n$ unrewarded edges are removed in $M_n$ in comparison to $M$. Note that the state space of $M_n$ remains identical to $M$ and that $\mathcal{D}_{M_n}=\frac{3}{101-n}$, therefore $\mathcal{D}_{M_n} < \mathcal{D}_{M_{n+1}}$.

We randomly explore the state space for 200k steps with a maximal time horizon $h$ of 12, 13 and 14 and record the exit rate at the end of the episode. We plot the mean exit rate according to the reward density in \autoref{fig:density} (left). We can see that the exit rate decreases when the reward sparsity increases, thereby illustrating that reward sparsity is not a reliable indicator of the difficulty of an MDP.

\section{Zero-Incentive Dynamics}
\label{sec:partially-ordered-tasks}
We argue that \citet{ocana_overview_2023} and \citet{molinghen_laser_2024} respectively describe the same phenomenon albeit with different terminology. The former use the term of Partially Ordered Subtasks (POS) to emphasize that there are subtasks from a high-level perspective, while the latter use the term of Zero-Incentive Dynamics (ZID) to emphasize the absence of reward for achieving these subtasks. In this work, we bring these two concepts together and use the term ``ZID'', arguing that the question of the reward is central, as we show further in this section.

In order to provide further insight into why ZID are challenging to deep RL, we first formally define ZIDs (equivalently POSs) from a graph-theoretical perspective, show the inability of state-of-the-art methods to cope with problems with ZID, and provide experimental evidence of the importance of rewarding subtasks shortly after their completion.

\subsection{Definitions}
\label{sec:zid-definitions}
\citet{molinghen_laser_2024} refer to the fact that agents must collaborate in LLE to block lasers and allow other agents to pass as \textit{agent interdependence} and argue that agent interdependence creates State Space Bottlenecks (SSB), although no formal definition is provided. 

Intuitively, in an MDP with a mandatory subtask, an SSB is the set of state-action pairs that directly lead to the accomplishment of this subtask. Formally, we define an SSB in the context of a graph induced by an MDP. For the sake of conciseness, we provide definitions of the concepts of graph theory required for \autoref{def:bottleneck} in \autoref{apx:graph-definitions}.

\begin{definition}[State Space Bottleneck]
    \label{def:bottleneck}  
    Let $M = \left<S, A, R, T\right>$ be an MDP and $G_M=(V, E, W)$ be the directed weighted graph induced by $M$. Let $S_0$ be the set of initial states and $S_G$ be the set of goal states.\\    
    A state space bottleneck $\mathcal{B}$ is a minimum directed $S_0$-$S_G$ cut-set of $G_M$.
\end{definition}

We proceed to build on top of SSBs to propose a formal definition of ZID. Intuitively, an SSB has ZID if completing the according subtask is not rewarded.

\begin{definition}[Zero-Incentive Dynamics]
    Let $M$ be an MDP and $G_M=\left(V, E, W\right)$ be the directed weighted graph induced by $M$. Let $\mathcal{B}$ be a bottleneck of $M$. Let $r_b$ be the base reward in $M$.\\
    $\mathcal{B}$ has zero-incentive dynamics if $\forall e \in \mathcal{B},~W(e) \leq r_b$.
\end{definition}

In some cases, it could be the case that the accomplishment of a subtask is not directly rewarded, for instance because a human has to validate the accomplishment or because of network latency. In those cases, one would rather talk about \textit{Delayed Incentive Dynamics} with some delay $d$.

\subsection{Subgoal-oriented RL under ZID}
\label{sec:subgoal-oriented}
There exist a wide literature on subgoal-oriented RL, and the question of whether state-of-the-art methods in subgoal-oriented RL are able to deal with ZID therefore naturally arises. As such, we evaluate MASER \citep{jeon_maser_2022} and HAVEN \citep{xu_haven_2023}, two state-of-the-art methods that aim at leveraging the presence of subgoals in the environment and that have shown to outperform other algorithms such as VDN \citep{sunehag_value-decomposition_2018} and QMIX \citep{rashid_qmix_2018} in the StarCraft Multi-Agent Challenge \citep{samvelyan_starcraft_2019}. Internally, MASER uses the expected return at both the individual and at the collective level as a metric to identify subgoals and design an intrinsic reward signal to encourage transitions that reach the identified subgoals. HAVEN uses a hierarchical structure (see \autoref{sec:hrl}) where the meta-agent uses its own policy to identify subgoals and assign them to the agents.

\begin{figure}
    \centering
    \includesvg[width=0.6\linewidth]{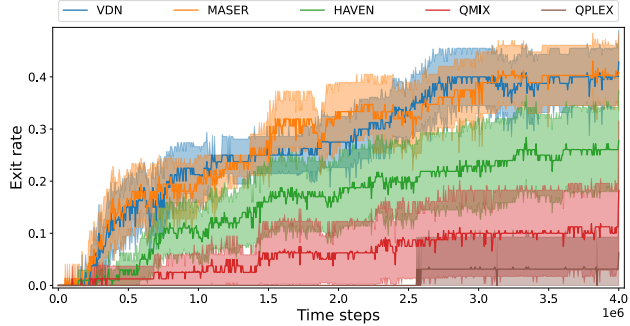}
    \caption{Average exit rate over the course of the training. Results are shown with 95\% confidence intervals and averaged over 16 different seeds. The maximal achievable exit rate is 1.}
    \label{fig:subgoal-oriented}
\end{figure}

We train VDN, QMIX, QPLEX \citep{wang_qplex_2021}, HAVEN and MASER agents (see \autoref{apx:haven-maser} for the full methodology) on the environment illustrated in \autoref{fig:lle} for which \citet{molinghen_laser_2024} have already shown that VDN and QMIX were unable to complete it. We plot the mean exit rate over the course of the training in \autoref{fig:subgoal-oriented} and observe that neither HAVEN nor MASER are able to outperform VDN in this setup, thereby showing the inability of these methods to identify -- or at least leverage the presence of -- subgoals in the environment. We hypothesize that due to their internal working principles, HAVEN and MASER both implicitly require subgoals to be rewarded to exploit their presence and are therefore unable to outperform other algorithms under ZID. We further motivate this hypothesis in \autoref{apx:haven-maser} by showing that removing ZID enables these methods to complete the task seamlessly.

\subsection{Impact of the SSB-to-reward distance}
\label{sec:experimental-investigation}
\begin{figure}
    \centering
    \begin{subfigure}[t]{0.65\linewidth}
        \includesvg[height=0.17\textheight]{plots/delay.svg}
    \end{subfigure}
    \begin{subfigure}[t]{0.24\linewidth}
        \centering
        \includegraphics[width=\textwidth]{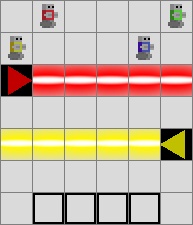}
    \end{subfigure}
    \caption{(Left) Exit rate over training time step for shaped reward delay $d=0, 1, \dots, 4$ on the environment shown on the right. Results are averaged on 30 different seeds and shown with 95\% confidence intervals. (Right) Environment used to analyse nuances of ZID. The agents randomly spawn in the two top rows and must reach the exits represented by the tiles with a black frame.}
    \label{fig:delayed-reward}
\end{figure}

To further understand the impact of ZID on deep RL, let us consider the environment shown in \autoref{fig:delayed-reward} (right) where the two lasers are SSBs with ZID. To show the effects of ZID on deep RL, we design a slightly different version of this environment in which, once per episode, a collective reward is received the first time each agent crosses each laser. To ensure that the optimal policy $\pi^*$ remains unchanged, we perform this reward shaping with PBRS as detailed in \autoref{apx:delayed-reward-methodology}.

We train agents with VDN and analyse the exit rate over the course of the training. Our full methodology can be found in \autoref{apx:delayed-reward-methodology}. In \autoref{fig:delayed-reward} the curve labelled ``No shaping'' (brown) is the exit rate for the initial problem with ZID, and the curve labelled ``$d=0$'' (blue) is the exit rate for the shaped version. The plot shows the blatant effects of ZID as the agents in the shaped version (blue) learn a good policy significantly faster than the others (brown).

We further investigate the effect of ZID by introducing a delay of $d$ steps between the accomplishment of a subtask and the collection of the corresponding reward. Our results in \autoref{fig:delayed-reward} clearly show that the closer the reward is to the accomplishment of the subtask, the better the policy. 

Hence, we argue that the delay between the accomplishment of a subtask and the collection of the according reward is a key factor to the learning of deep RL policies. This result also further disqualifies reward density as an indicator of the problem difficulty since the reward density is identical for $d = 0,\dots, 4$ but the policy is different.

\section{Discussion}
\label{sec:discussion}
On the one hand, the negative results of \autoref{sec:subgoal-oriented} show that subtasks with ZID remain a challenge to state-of-the-art deep RL methods to this day. On the other hand, the positive results of \autoref{sec:experimental-investigation} suggest that if agents were able to automatically identify the accomplishment of subtasks, even a few steps late, it would be readily possible to mitigate or even overcome ZID.

\citet{sunel_faster_2024} discuss three categories of automatic subgoal identification methods: graph-based approaches that build a (sub-)graph of the environment and use algorithms on graphs to identify useful edges or vertices; statistics-based approaches that compute metrics on states or transitions; and Multiple Instance Learning (MIL) based approaches that classify episodes as either positive (task success) or negative (task failure) and typically identify common features among positive instances. We readily disqualify graph-based approaches such as the one of \citet{simsek_identifying_2005} on the grounds that their ability to identify subgoals scales poorly with the size of the state space as illustrated in \autoref{apx:graph-based-approaches}. MIL methods \citep{mcgovern_automatic_2001} are not readily applicable to our problem because they rely on the comparison between positive (success) and negative (failure) trajectories. As shown in \autoref{sec:subgoal-oriented}, SOTA algorithms are unable to complete ZID problems, which makes it impossible to gather positive instances and apply MIL methods. Statistics-based approaches include a wide variety of methods that assign a metric to each state and leverage that metric to identify subgoals. MASER and HAVEN are examples of such methods and have shown to be ineffective.

\section{Conclusion}
In this paper, we have first shown that reward sparsity is a topic more complex than it looks. Specifically, we showed that increasing the reward density of a task could counter-intuitively lower its success rate, thereby showing that reward sparsity is not a reliable indicator of the difficulty of a problem. Then, we specifically focused on Zero-Incentive Dynamics, a particular case of reward sparsity where the accomplishment of subgoal is not rewarded. We defined the concept formally, and then showed that state-of-the-art subgoal-oriented methods are unable to identify subgoals under Zero-Incentive Dynamics and therefore performed as poorly as more general-purpose deep RL methods. To give more insights on the importance of rewarding subtasks, we showed that the distance between the achievement of a subtask and the collection of the corresponding reward is a key factor to the quality of the learned policy, thereby raising the question of whether subtasks could be discovered automatically. We discuss this possibility and argue that there exist no such method to this day.

Together, our results shed light on a specific kind of reward sparsity -- Zero-Incentive Dynamics -- and show that there exist no counter-measure to this day, thereby indicating a path for future research.

\appendix
\renewcommand{\sectionautorefname}{Appendix}

\section{Graphs and cuts}
\label{apx:graph-definitions}
The following definitions of graph theory \citep{wilson_introduction_2009} are used to formalize the concepts of State Space Bottleneck and Zero-Incentive Dynamics in \autoref{sec:zid-definitions}.

\begin{definition}[Directed cut]
    Let $G = (V, E)$ be a directed graph. A directed cut $\left<S, T\right>$ (or dicut) of $G$ is a partition of $V$ into two disjoint sets $S$ and $T$, such that each edge is directed from $S$ to $T$.
\end{definition}

\begin{definition}[Cut-set]
    A cut $\left<S, T\right>$ defines a cut-set $C$, the set of edges that have one endpoint in $S$ and one endpoint in $T$.
\end{definition}

\begin{definition}[Minimum cut]
    A cut is minimum if the size of the cut-set is not larger than the size of any other cut-set.
\end{definition}

\begin{definition}[$S$-$T$ cut]
    Let $G = (V, E)$ be a graph.\\
    Let $S$ and $T$ be two disjoint subsets of $V$, i.e. $S \subset V$, $T \subset V$ and $S \cap T = \emptyset$.

    An $S$-$T$ cut is a cut $\left<A, B\right>$ of $G$ such that $\forall s \in S, s \in A$ and $\forall t \in T, t \in B$.
\end{definition}

\begin{definition}[Winning walk]
    A winning walk is a finite sequence of pairwise adjacent edges $e_1, e_2, \dots, e_n$ such that $e_1 = \left<s_0, a, s'\right>$ and $e_n = \left<s, a, s_g\right>$ with $s_0 \in S_0$ and $s_g \in S_G$.
\end{definition}

\section{Subgoal-oriented approaches methodology}
\label{apx:haven-maser}
We train the agents with VDN, QMIX, QPLEX, HAVEN and MASER on the map shown in \autoref{fig:lle} for 4 million time steps with double deep $Q$-learning \citep{van_hasselt_deep_2016}, an $\epsilon$-greedy exploration policy where $\epsilon$ is annealed from $1$ down to $0.05$ over 50k time steps, and set a maximal episode time limit to $\lfloor{\frac{width \times height}{2}\rceil} = 78$ steps. The $Q$-network is a Convolutional Neural Network \citep[CNN]{lecun_gradient-based_1998} of three layers (with 32, 64 and 32 filters respectively and a kernel of size 3) that are flattened and then fed through three linear layers of 64 neurons. All layers have a ReLU activation function. Since the agents share the same parameters, we concatenate the flattened output of the CNN with their one-hot encoded agent ID. We optimize the $Q$-network and the mixer after every episode on a batch of 32 episodes with an ADAM optimizer for both the $Q$-network and the mixer with a learning rate of 5$\times$10\textsuperscript{-4}, use a $\gamma=0.95$, clip the norm of the gradients to 10, update the target network every 200 time steps, and use a replay memory of 5k episodes. For HAVEN, we use VDN as mixing network and the meta-agent time scale $k=3$. For MASER, the intrinsic reward weight $\lambda=0.03$, the weight of individual vs global targets is 0.5 and we use VDN as mixing method.

\begin{figure}
    \centering
    \includesvg[width=0.8\linewidth]{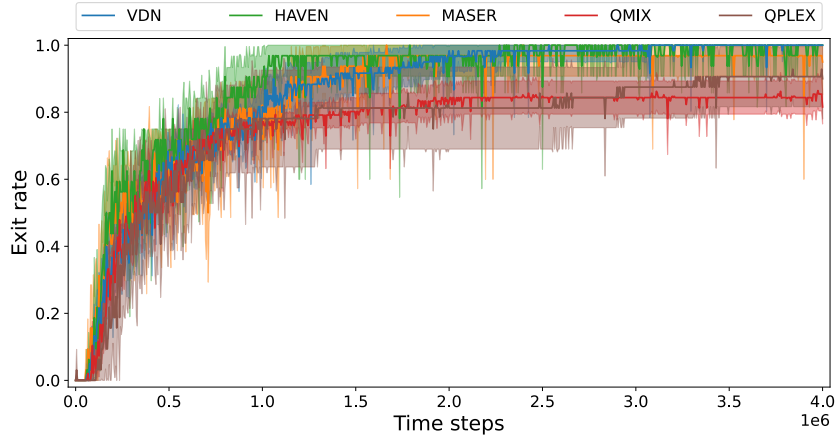}
    \caption{Exit rate over the course of the training for subgoal-oriented methods when PBRS is applied. Results are averaged on 16 different seeds and shown with 95\% confidence intervals. All algorithms successfully complete the collaborative task (i.e. an exit rate of 1).}
    \label{fig:apx:subgoals-with-shaping}
\end{figure}

Additionally, we provide in \autoref{fig:apx:subgoals-with-shaping} the results with the same methodology but when the completion of subtasks is rewarded, which shows that all methods successfully complete the collaborative task when PBRS is applied as described in \autoref{apx:delayed-reward-methodology} with $d=0$.

\section{Delayed PBRS methodology}
\label{apx:delayed-reward-methodology}
Let $C$ be 2-dimensional a matrix initialized at $-1$ that indicates at $C_{i,l}$ since how many steps agent $i$ has crossed laser $l$ during the current episode. We define the potential function $\phi$ as shown in \autoref{eq:delayed-potential-shaping}.

\begin{equation}
    \label{eq:delayed-potential-shaping}
    \phi(s) = -\sum_{i=1}^{n}\sum_{l} \begin{cases}
        1 \text{~if~} C_{i, l} \leq d\\
        0 \text{~otherwise}
    \end{cases} 
\end{equation}

In order to let agents discriminate between a state where a shaped reward has already been received for blocking a specific laser or not, we concatenate the inputs of agent $i$ with $C_i$, i.e. the value of its corresponding countdowns.

We train agents with VDN \citep{sunehag_value-decomposition_2018} for 300k time steps on the LLE map shown in \autoref{fig:delayed-reward} where the agents randomly spawn above the top laser. We use an $\epsilon$-greedy policy where $\epsilon$ is annealed from 1 down to 0.05 over 100k time steps, the maximal episode time limit is set to 28 steps and the replay memory stores up to 50k transitions. The $Q$-network is identical to the one described in \autoref{apx:haven-maser}. We optimize the $Q$-network and the mixer every 5 steps on a batch of 64 transitions with an ADAM optimizer with a learning rate of 5$\times$10\textsuperscript{-4}, clip the norm of the gradients to 10 and update the target network every 200 time steps.
Since the agents share the same parameters, we concatenate the flattened output of the CNN with their one-hot encoded agent ID and with their own row $C_i$.

Note that when an episode is truncated because the time horizon has been reached, the pending rewards are flushed in order to keep the same reward density across all values of $d$.

%%%%%%%%%%%%%%%%%%%%%%%%%%%%%%%%%%%%%%%%%%%%%%%%%%%%%%%%%%%%%%%%
%% Bibliography
%%%%%%%%%%%%%%%%%%%%%%%%%%%%%%%%%%%%%%%%%%%%%%%%%%%%%%%%%%%%%%%%
\bibliography{references}
\bibliographystyle{rlj}

%%%%%%%%%%%%%%%%%%%%%%%%%%%%%%%%%%%%%%%%%%%%%%%%%%%%%%%%%%%%%%%%
% AUTHOR: If your paper has no supplementary materials, you may 
%         comment out the line below, which creates the title for
%         the supplementary materials.
%%%%%%%%%%%%%%%%%%%%%%%%%%%%%%%%%%%%%%%%%%%%%%%%%%%%%%%%%%%%%%%%
\beginSupplementaryMaterials

\section{Reward density computation}
\label{apx:density-computation}
The number of edges of \autoref{fig:density} is computed as follows.
\begin{itemize}
    \item We consider that bidirectional edges account for two edges;
    \item There are 5 actions (north, east, south, west and stay);
    \item The three centre states each have 5 neighbours, i.e. $9 \times 5= 45$;
    \item The four corners have 3 neighbours, i.e. $4 \times 3 = 12$;
    \item The exit tile has no neighbour, it is a goal state;
    \item The 11 remaining tiles on the border each have 4 neighbours, i.e. $11 \times 4 = 44$.
\end{itemize}

The total number of edges is 45 + 44 + 12 = 101 edges. There are three rewarded edges, that are the ones leading to the goal state. The reward density is therefore $\frac{3}{101}$.

\section{Graph-based approaches}
\label{apx:graph-based-approaches}
\begin{figure}
    \centering
    \begin{subfigure}{0.495\linewidth}
        \includegraphics[width=\linewidth]{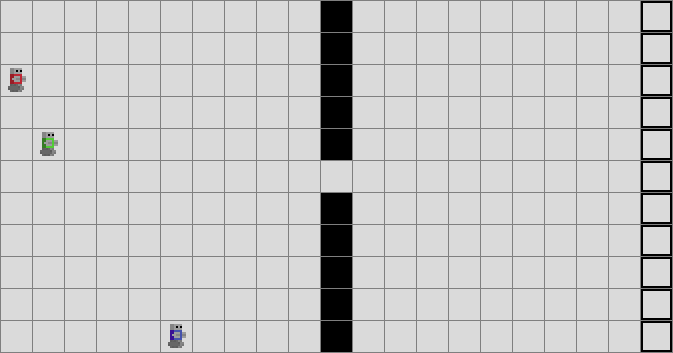}
        \subcaption{}
        \vspace{0.3cm}
        \label{subfig:3agents}
    \end{subfigure}
    \begin{subfigure}{0.495\linewidth}
        \centering
        \includesvg[width=\linewidth]{pictures/1-agents-bottleneck.svg}
        \subcaption{}
        \vspace{0.3cm}
    \end{subfigure}
    \begin{subfigure}{0.495\linewidth}
        \includesvg[width=\linewidth]{pictures/2-agents-bottleneck.svg}
        \subcaption{}
        \vspace{0.3cm}
    \end{subfigure}
    \begin{subfigure}{0.495\linewidth}
        \includesvg[width=\linewidth]{pictures/3-agents-bottleneck.svg}
        \subcaption{}
        \vspace{0.3cm}
    \end{subfigure}
    \begin{subfigure}{\linewidth}
        \includesvg[width=\linewidth]{pictures/colourbar-horizontal.svg}
    \end{subfigure}
    \caption{(a) Map with a State Space Bottleneck. (b), (c) and (d) respectively show the bottleneck score for each vertex to be identified as a bottleneck for one, two and three agents. Bright colours indicated a high score while dark colours indicates a low score.}
    \label{fig:automatic-discovery}
    \vspace{0.7cm}
\end{figure}

There are multiple graph-based approaches, and we investigate the one of \citet{simsek_identifying_2005} who build an approximation of the MDP during training and then use spectral clustering to identify bridges in the graph, i.e. SSBs. We adapt this method to the multi-agent case and test it on the very simple map shown in \autoref{subfig:3agents} that has one explicit SSB and matches the one of the original work.

\paragraph{Methodology} We record training episodes and after every 5 episodes, we build a connected local graph of the MDP according to the encountered states and edges. Then, we perform a binary spectral clustering \citep{shi_normalized_2000} of vertices to identify edges that lie at the intersection of densely connected areas of the state space. Over the course of the training, agents build a database of how many times each edge has been identified as a bottleneck which we refer to as its score.

Similarly to \citet{simsek_identifying_2005}, we represent the results on the vertex level rather than on the edge level for visualization purposes. To do so, we assign to each vertex $v = (x, y)$ the sum of the scores of all the edges where at least one agent is in $(x, y)$, as shown in \autoref{eq:vertex-score} where $(x_i, y_i)$ is the location of agent $i$ in a given state $s$.

\begin{equation}
    \label{eq:vertex-score}
    \text{score}_{x,y} = \sum_{s \in S} \sum_{i=1}^n\begin{cases}
        \text{score}(s) \text{ if } (x_i, y_i) = (x, y)\\
        0 \text{ otherwise}
    \end{cases}
\end{equation}

\begin{figure}
    \centering
    \includesvg[width=0.7\linewidth]{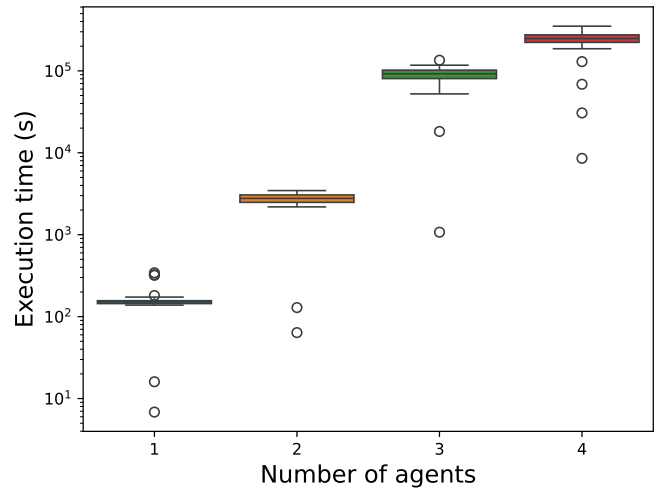}
    \caption{Average execution time (in seconds, logarithmic scale) to identify bottlenecks in the map shown in \autoref{subfig:3agents} for one to four agents. Results are averaged on 30 runs.}
    \label{fig:execution-time}
\end{figure}

We randomly explore the state space for 1 million time steps with one to four agents to identify subgoals in the map shown in \autoref{subfig:3agents}. We show the subgoal identification for one to three agents in \autoref{fig:automatic-discovery}, and the corresponding computation duration in \autoref{fig:execution-time}.

\paragraph{Results}

As shown in \autoref{fig:automatic-discovery}, although the bottleneck identification works very well for a single agent, the bottleneck identification decreases with the number of agents. With two agents, the bottleneck can still be identified but with much less confidence, as indicated by the slightly lighter colour of the bottleneck state, but from three agents onwards, the bottleneck state is no longer identified. Moreover, the computation time exponentially increases with the number of agents, as shown in \autoref{fig:execution-time}, which makes this method unusable in practice.

We conclude that this method scales poorly with the number of agents in terms of ability to identify subgoals and in terms of computation time. We attribute this phenomenon to the low connectivity of the sub-graphs because of the exponential growth the the state space with the number of agents that make it unlikely to encounter the same state multiple times on the course of an episode. We expect analogous methods \citep{kazemitabar_strongly_2009} to face similar challenges.
\end{document}